\RequirePackage{fix-cm}

\documentclass[twocolumn]{svjour3}          % twocolumn

\smartqed  
\usepackage{graphicx}
\usepackage{amsfonts}
\usepackage{amsmath}
\usepackage{latexsym}

\journalname{arXiv Preprint}
\begin{document}

\title{Validating GAN-BioBERT: A Methodology For Assessing Reporting Trends In Clinical Trials \thanks{This study was funded by the University of Wisconsin School of Medicine and Public Health's Shapiro Summer Research Program.} }

\titlerunning{Validating GAN-BioBERT}        

\author{Joshua J Myszewski, BSE         \and
       	Emily Klossowski, MS 		\and
	Patrick Meyer, MD 		\and
	Kristin Bevil, MD 		\and
	Lisa Klesius, MD 	\and
	Kristopher M Schroeder, MD %etc.
}

\institute{Joshua Myszewski \at
              University of Wisconsin School of Medicine and Public Health, 600 Highland Ave, B6/319 Madison, WI, 53792, USA \\
              Tel.: +1262-391-8003\\              
              \email{jmyszewski@wisc.edu}           %  \\
           \and
           Kristopher M Schroeder \at
              University of Wisconsin School of Medicine and Public Health Department of Anesthesiology, 600 Highland Ave, B6/319 Madison, WI, 53792, USA
}

\date{Received: 06/01/2021}

\maketitle

\begin{abstract}
\hfill\break
{\bf Purpose:} In the past decade there has been much discussion about the issue of biased reporting in the context of clinical research. Despite this attention, there have been limited tools developed for the systematic assessment of qualitative statements made in clinical research, with the majority of studies assessing qualitative statements relying on the use of manual expert raters, which limits the ultimate size of the studies. Additionally, previous attempts to develop larger scale tools, such as those using natural language processing, were limited by both their accuracy and the number of categories used for the classification of their findings. With these limitations in mind, the goal of this study was to develop a classification algorithm that was both suitably accurate and finely grained to be applied on a large scale for the assessment of the qualitative sentiment expressed in clinical trial abstracts. Additionally, this study seeks to compare the performance of the proposed algorithm, GAN-BioBERT, to previous studies as well as to expert manual rating of clinical trial abstracts. 
{\bf Methods:} This study develops a three-class sentiment classification algorithm for clinical trial abstracts using a semi-supervised natural language process model based on the Bidirectional Encoder Representation from Transformers (BERT) model, from a series of clinical trial abstracts annotated by a group of experts in academic medicine.
{\bf Results:} 
The use of this algorithm was found to have a classification accuracy of 91.3\%, with a macro F1-Score of 0.92, which is a significant improvement in accuracy when compared to both previous methods and expert ratings, while also making the sentiment classification finer grained than previous studies. 
{\bf Conclusion:} The proposed algorithm, GAN-BioBERT, is a suitable classification model for the large-scale assessment of qualitative statements in clinical trial literature, providing an accurate, reproducible tool for the large scale study of clinical publication trends.

\end{abstract}

\section{Introduction}
\label{intro}
Publication bias is a widespread phenomenon of systematic under or overreporting of research findings dependent on the direction of the results found \cite{McGauran}. As a result of this phenomenon, systematic reviews of clinical guidelines may reach incorrect conclusions \cite{Sutton}, which creates the potential for harm to patients with treatments that have an otherwise poor evidence base. Despite this potential for harm and its widespread presence within clinical literature \cite{McGauran}, there have been limited efforts to develop and utilize methods to characterize publication bias on a large scale.
In 2016 Hedin et. al. found that only fifty-five percent of meta-analysis in anesthesiology journals discussed publication bias, and only forty-three percent actually used tools to assess the phenomenon \cite{Hedin}. Furthermore, currently used methods for assessing publication bias such as funnel-plot based methods and selection models \cite{Lin,Eggers}, are criticized as unintuitive to interpret within the literature's context \cite{Lin}. These methods are also focused on the quantitative findings expressed in the studies in question in the form of effect sizes and p-values, and are therefore limited to those studies that express these types of findings. 

Historically, systematic assessment of the qualitative interpretation of the findings has been limited to rating systems performed by human raters \cite{Oliveira,Yuan,Vecchi}. This has changed with the development of sentiment analysis and natural language processing as a toolset capable of understanding the qualitative statements made in a body of text. Several studies have explored the assessment of citation sentiment analysis in academic literature \cite{Xu,Aljuaid,Yousif,Kilicoglu} with the goal of examining the sentiment towards particular papers cited in the body of another article as an assessment of article impact. Sentiment analysis has also been applied to the analysis of clinical notes in the electronic health record with the goal of prognostication \cite{Weissman,Ghassemi}. In contrast, there have been limited attempts to use sentiment analysis to characterize the qualitative findings authors express towards their own publication's findings \cite{Zlabinger,Fischer}. However, both of these studies were limited by the availability of labeled abstract data for training of the algorithms developed, as well as the algorithms classes being limited to the two class tasks of positive/neutral \cite{Zlabinger} or positive/not positive \cite{Fischer} respectively. The methods used in these studies also did not take advantage of newer natural language processing architectures such as the Bidirectional Encoder Representations from Transformers (BERT) model \cite{BERT}, or similar newer models for the analyses, instead opting for the use of a support vector machine \cite{Zlabinger} and a sequential neural network \cite{Fischer} respectively. 

This study's goal is to develop and validate a sentiment analysis model for clinical trial abstracts that addresses the limitations of previous studies such that it can be applied to large-scale assessment of clinical literature. This was done by fine tuning a BERT model for three-class sentiment classification \cite{BERT} pretrained on PubMed text \cite{BioBERT} in a semi-supervised fashion with the assistance of a generative adversarial network (GAN) \cite{GANBERT}. Training was performed using clinical trial abstracts with sentiment labeled by expert raters supplemented with a large amount of unlabeled abstracts from the PubMed database. The final performance of this algorithm was then determined and compared to previous methodologies used for this type of sentiment analysis task.

\section{Methods}
The overall goal of this study was the creation of a semi-supervised three-class sentiment classifier for clinical trial abstracts that classifies clinical trial abstracts as expressing positive, negative, or neutral sentiment. This approach was chosen so that the final algorithm could be both sufficiently accurate for application in future studies, while also providing enough information to delineate publication trends in clinical trial literature over time. 

\begin{figure*}[htb]

  \fbox{\includegraphics[width=0.75\textwidth]{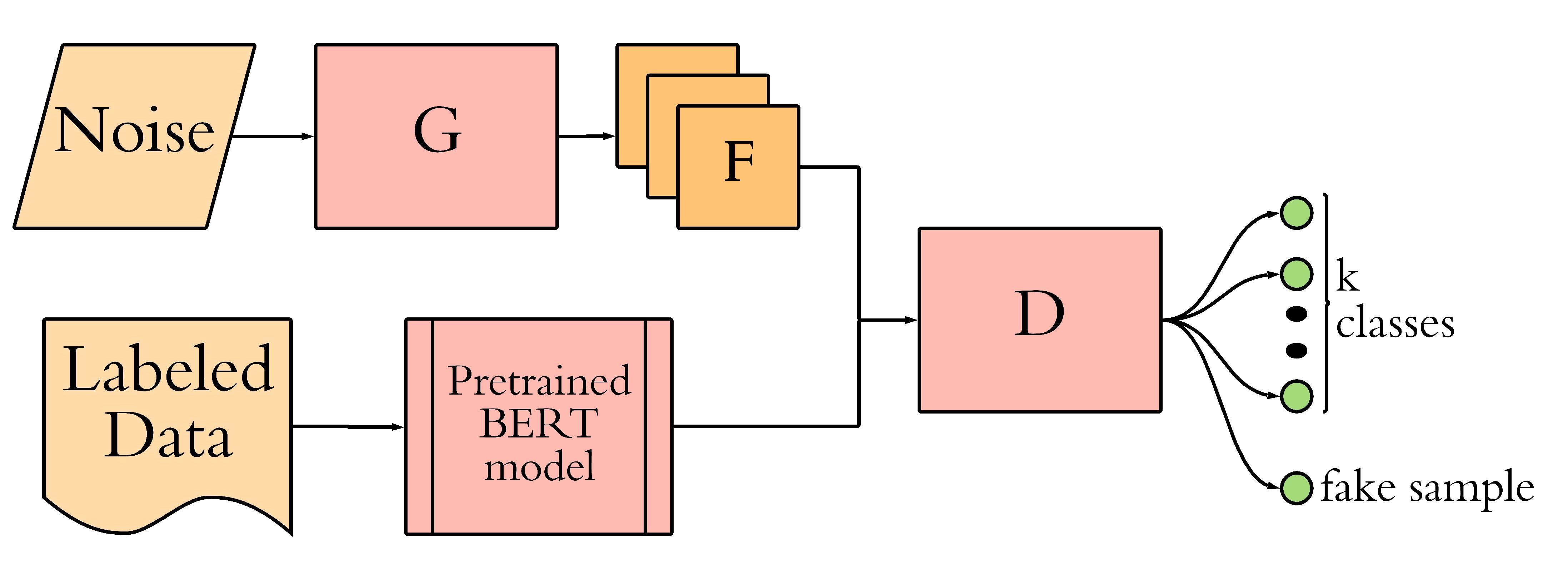}}

\caption{A visual representation of the GAN-BERT algorithm as described by the original developers \cite{GANBERT} }
\label{fig:1}      
\end{figure*}

\subsection{Creation of the Labeled \& Unlabeled Training Sets} 
There are no publicly available annotated datasets specific to sentiment analysis of clinical trials, so for the purposes of this study an appropriate annotated dataset had to be created. Given that the most appropriate raters for the sentiment rating of clinical trials are trained clinician experts, creation of a fully annotated dataset for this study was determined to be particularly resource intensive, a problem inherent to clinically related natural language processing (NLP) tasks \cite{Xia}. With this in mind, this study elected to use a semi-supervised approach, combining expert-annotated clinical trial abstracts and large amounts of unlabeled data to minimize the resources required to create the final algorithm. 

\paragraph{Data Gathering}All abstracts gathered for this study were from the National Library of Medicine's (NLM) PubMed database, filtered specifically to publications classified as clinical trials. The collection of these abstracts was automated using the NCBI's Entrez search and retrieval system with a data mining tool built by the authors using the BioPython toolkit \cite{Biopython}. This tool is able to gather all MEDLINE data that is reported for a particular PubMed query, and is able to search in a specific medical field by cross referencing journal ID numbers with a NLM catalog query. 
 
For the creation of the labeled dataset, 12 abstracts each from clinical trials in the fields of Obstetrics \& Gynecology, Orthopedics, Pediatrics, Anesthesiology, General Surgery, Internal Medicine, Thoracic Surgery, Critical Care, and Cardiology were randomly selected, for a total of 108 labeled abstracts. These abstracts were then stripped of all information other than the abstract text, and provided to a panel of four clinician experts to independently label the abstracts as having positive, negative, or neutral sentiment. The ground truth class of these abstracts was then defined by the most common rating assigned to the abstract by three of the raters. A single rater's scores were held out to serve as an accuracy measure to compare with the validation results of the final algorithm. The smallest and largest class of the labeled data was then randomly oversampled or undersampled to equal the number of samples from the median class in order to create an algorithm with a maximally balanced accuracy between classes \cite{Wei}.
The unlabeled dataset was a collection of 2000 clinical trial abstracts selected from PubMed in the same manner described above, excluding those used in the labeled dataset. The unlabeled data is then given a label of UNK\_UNK so that when it's used to train the classification algorithm the label is appropriately masked.

\paragraph{Data Preprocessing} The conclusion sentences of the labeled and unlabeled abstracts to be used for training and validation were then extracted. This was done as it was found in previous study that using solely the concluding sentences led to an increase in classification accuracy \cite{Zlabinger,Fischer}. Using the Natural Language Tool Kit (NLTK) Python toolkit \cite{NLTK}, concluding sentences were identified as those following the conclusion heading for structured abstracts. For unstructured abstracts, the conclusion sentences were determined to be the last n sentences of an abstract based on the number of sentences in the abstract using equation 1 below, where S\textsubscript{t} is the total number of sentences. 
\begin{equation}
n = max( S_t * 0.125)
\end{equation}
The relative value of 0.125 was determined empirically in a previous study based on the analysis of 2000 structured abstracts \cite{Zlabinger}. 

\paragraph{Tokenization} Following extraction of the conclusion sentences, all sentences were tokenized using the BERT tokenizer available as part of the HuggingFace Transformers toolkit \cite{HuggingFace}, which tokenizes each word. The BERT tokenizer begins by tagging the first token of each sentence with the token [CLS], then converting each token to its corresponding ID that is defined in the pre-trained BERT model. The end of each sentence is then padded with the tag [PAD] to a fixed sentence length, as the BERT model requires a fixed length sentence as an input \cite{BERT}.

\subsection{GAN-BioBERT Workflow}
Generally, the GAN-BERT architecture consists of a generator function G based on the Semi-Supervised generalized adversarial network (GAN) architecture that generates fake samples F using a noise vector as input \cite{SSGAN}, the pre-trained BERT model, which is given the labeled data, and a discriminator function D that is a BERT-based k-class classifier that is fine-tuned to the particular classification task \cite{GANBERT}. This workflow is shown graphically in Figure~\ref{fig:1}, with further discussion of each element to follow. 

\subsection{BERT Architecture}
Before discussing the details of the algorithm used in this study it's key to first discuss the general BERT architecture. Bidirectional Encoder Representations from Transformers or BERT model is a method for language processing first described in 2018 by Devlin et. al. that achieved state of the art performance on a variety of natural language processing tasks and has since become a heavily used tool in natural language processing research \cite{BERT}. BERT functions using 2 sequential workflows, a semi-supervised language modeling task that develops a general language model, then a supervised learning step specific to the language processing task the model is being applied to such as text classification. For developing the pre-trained language model BERT is provided with a very large corpus from a particular domain, such as publications in PubMed \cite{BioBERT}, documents from a particular language \cite{LanguageBERT}, or English Wikipedia and BooksCorpus as in the original BERT model \cite{BERT}. BERT then develops a complete language model from the provided corpus using both masked language modeling, which determine the meaning of individual words within the sentence's context, and next sentence prediction, which works to understand the relationship between sentences. The result of this process is a trained context-sensitive general language model for the specific domain being studied that can then be disseminated for a wide variety of applications. The pretrained language model from the semi-supervised stage of BERT is then fine-tuned for a specific language task by providing task-specific inputs and outputs and then adjusting the parameters of the model accordingly to create the complete task-specific algorithm \cite{BERT}. 

\paragraph{BERT Pretrained Model Selection} 
Given the important role of the pretrained model in the BERT architecture, and the relative complexity of biomedical literature, general language models are likely to encounter lower accuracy when applied to a biomedical application such as the one in this study due to a change in the word distributions between general and biomedical corpora \cite{BioBERT}. As such, in this study the pretrained BioBERT model was used as the general language model to be fine-tuned for sentiment classification \cite{BioBERT}. BioBERT is a 2020 pretrained BERT model by Lee et. al. that is specific to the biomedical domain that was trained on PubMed abstracts and PubMed Central full-text articles, as well as English Wikipedia and BooksCorpus as was done in the original BERT model \cite{BERT,BioBERT}. As a result of this domain specific training, BioBERT has shown improved performance on a variety of biomedical NLP tasks when compared to the standard BERT models \cite{BioBERT}. 

\subsection{GAN-BERT}
While BERT and its derivatives have been able to achieve state of the art performance on a variety of tasks, one major limitation of the model is that fully trained models typically require thousands of annotated examples to achieve these results \cite{GANBERT}. In particular, significant drops in performance were observed when less than 200 annotated examples are used \cite{GANBERT}. In order to address this limitation, Croce, Castellucci, and Basili developed the GAN-BERT model in 2020 as a semi-supervised approach to fine tuning BERT models that achieves performance competitive with fully supervised settings \cite{GANBERT}. Specifically, GAN-BERT expands upon the BERT architecture by the introduction of a Semi-Supervised Generative Adversarial Network (SS-GAN) to the fine-tuning step of the BERT architecture \cite{SSGAN}. In a SS-GAN, a "generator" is trained to produce samples resembling the data distribution of the training data i.e. the labeled abstracts in this study. This process is dependent on a "discriminator", a BERT-based classifier in the case of this study, which in an SS-GAN is trained to classify the data into their true classes, in addition to identifying whether the sample was created by the generator or not. When trained in this manner, the labeled abstract data was used to train the discriminator, while both the unlabeled abstracts and the generated data is used to improve the model's inner representations of the classes, which subsequently increases the model's generalizability to new data \cite{GANBERT}. As a result of this approach the minimum number of annotated samples to train a BERT model is reduced from thousands, to a few dozen \cite{GANBERT}. Because of this effect, this study uses GAN-BERT to minimize the resource intensive process of creating an expert-annotated corpus of clinical trial abstracts. 

\subsection{GAN-BioBERT Architecture}
Having now discussed the individual elements of the GAN-BioBERT workflow and the reasoning for their use in this study, the specific methods and parameters used in this study, including the details of the GAN-BERT architecture used, will be described in greater detail. The generator, G, used is a Multi Layer Perceptron (MLP) that takes a 100 dimension noise vector drawn from the normal distribution N($\mu$, $\sigma^2$) and produces a fake vector \emph{h}\textsubscript{fake} $\in$  R\textsuperscript{D} of length 256. Similarly, the discriminator, D, is a MLP that takes as an input a vector $h_* \in R^d$ which can be either the \emph{h}\textsubscript{fake} produced by the generator, or \emph{h}\textsubscript{real}, for the real labeled and unlabeled examples. The last layer of D is a softmax-activated layer, whose output is a vector of logits corresponding to the categories positive, negative, neutral or fake. For forward propagation through the discriminator,  it should classify $h_*$ into the categories positive, negative, or neutral if $h_*$ = h\textsubscript{real} or the category fake if $h_*$ = h\textsubscript{fake}. During the training process, the goal is to optimize the two competing losses \emph{L}\textsubscript{D} and \emph{L}\textsubscript{G} corresponding to the loss functions for the discriminator and generator respectively. These loss functions are described more formally by Salimans et al. \cite{SSGAN} as follows: 

Given that $p_m(\hat{y} = y | x,y = real)$ is the probability provided by the model that a sample x is considered a real example and that $p_m (\hat{y} = y | x,y = fake)$ is the probability given by the model that a sample x is considered a fake (generated) example, the loss function  \emph{L}\textsubscript{D} of the discriminator is defined as:
\begin{equation}
\emph{L}\textsubscript{D} = \emph{L}\textsubscript{D\textsubscript{sup}} + \emph{L}\textsubscript{D\textsubscript{unsup}}
\end{equation}
Where \emph{L}\textsubscript{D\textsubscript{sup}} is the error in assigning the wrong class to a real example defined as:
\begin{equation}
\emph{L}\textsubscript{D\textsubscript{sup}} = -\mathbb{E}_{x,y\sim p_d} log[p_m(\hat{y} = y | x,y = real)]
\end{equation}

 and \emph{L}\textsubscript{D\textsubscript{unsup}} is the error assigned to incorrectly assigning an unlabeled examples as fake or failing to recognize a fake example defined as:
\begin{equation}
\begin{split}
\emph{L}\textsubscript{D\textsubscript{unsup}} =-\mathbb{E}_{x\sim p_d} log[1 - p_m (\hat{y} = y | x,y = fake)] \\
- \mathbb{E}_{x\sim G} log[1 - p_m (\hat{y} = y | x,y = fake)]
\end{split}
\end{equation}

Meanwhile, the loss function \emph{L}\textsubscript{G} of the generator is defined as:
\begin{equation}
\emph{L}\textsubscript{G} = \emph{L}\textsubscript{G \textsubscript{feature matching}} + \emph{L}\textsubscript{G \textsubscript{unsup}}
\end{equation}
Where \emph{L}\textsubscript{G \textsubscript{feature matching}} is the feature matching loss, which leads the generator to produce examples that are increasingly similar to real inputs. If f(x) is the activation on an intermediate layer of discriminator D, This loss is further defined as:
\begin{equation}
  \emph{L}\textsubscript{G \textsubscript{feature matching}} = ||\mathbb{E}_{x\sim P_d}f(x) - \mathbb{E}_{x\sim G} f(x) ||_2^2
\end{equation}
and \emph{L}\textsubscript{G \textsubscript{unsup}} is the error induced by fake examples being correctly identified by the discriminator, i.e. it penalizes the generator for the discriminator succeeding in identifying a fake example. This is further defined as: 
\begin{equation}
\emph{L}\textsubscript{G \textsubscript{unsup}} = \mathbb{E}_{x\sim G} log[1 - p_m (\hat{y} = y | x,y = fake)]
\end{equation}

\begin{table*}[t]
\centering

\small
\setlength\tabcolsep{2pt}

\caption{Performance metric results for both this study \& previous studies}
\label{tab:1}    
\resizebox{\textwidth}{!}{
\begin{tabular}{lllll}
\hline\noalign{\smallskip}
Study & Classification Method & Classification Type (class \#) & Accuracy & F1-Score  \\
\noalign{\smallskip}\hline\noalign{\smallskip}
This Study, 2021 & Expert Rater & Pos, Neg, Neutral (3) & 62\% & 0.603 \\
Fischer \&  Steiger, 2020 \cite{Fischer} & Word Frequency + Sequential Neural Network & Pos/Not Pos (2) & 73\% & Not Reported \\
Zlabinger et al., 2018 \cite{Zlabinger} & Uni-gram Features + Support Vector Machine (SVM) & Positive/Neutral (2) & 76\% & 0.72 \\
This Study, 2021 & GAN-BERT & Pos, Neg, Neutral (3) & 82.6\% & 0.824 \\
This Study, 2021 & GAN-BioBERT & Pos, Neg, Neutral (3) & 91.3\% & 0.92 \\
\noalign{\smallskip}\hline
\end{tabular}
}
\end{table*}

During back-propagation, the unlabeled examples only contribute to \emph{L}\textsubscript{D\textsubscript{unsup}}, such that they only contribute to the loss if they are falsely considered real, otherwise, their contribution is masked \cite{GANBERT}. In contrast, the labeled examples contribute to the supervised loss \emph{L}\textsubscript{D\textsubscript{sup}} \cite{GANBERT}. It is important to note that in the GAN-BERT architecture the generated examples contribute to both the generator and discriminator loss \cite{GANBERT}. During back propagation, both the BERT weights and the discriminator's weights are updated, thus both the labeled and unlabeled data contribute to the final classification algorithm. After training the generator is discarded, leaving the BERT model for inference, with no additional computational cost when compared to using a standard BERT model \cite{GANBERT}.

In short, in this study GAN-BioBERT takes the BERT architecture pretrained on biomedical text using BioBERT \cite{BioBERT}, and fine-tunes it for sentiment classification of clinical trial abstracts in a semi-supervised manner by using adversarial learning in the form of an SS-GAN architecture known as GAN-BERT \cite{GANBERT,SSGAN}. The training data used consisted of a set of clinical trial abstracts annotated by three expert raters as positive, negative, or neutral, where the least common class was upsampled and the most common class was downsampled in order to create a balanced training set, as well as 2000 unlabeled clinical trial abstracts. The validation accuracy and F1-scores of the resulting algorithm were then determined and compared to previous attempts at this type of application, the performance of a fourth expert rater on the same labeled data used to train and validate the algorithm, and performance using the GAN-BERT algorithm without BioBERT (i.e. with the standard BERT pretrained model).

\section{Results and Discussion}

Of the 108 abstracts labeled by the expert raters, twenty-six were classified as positive, sixty-nine were classified as neutral, and thirteen were classified as negative by the raters. As such, the negative samples were upsampled, and the neutral samples were downsampled so that each class contained twenty-six examples, for a final labeled dataset of seventy-eight abstracts for training purposes. From this 30\% of the samples were held out as the validation set for determining the performance of the algorithm. 

\begin{figure}[h]

  \fbox{\includegraphics[width=0.45\textwidth]{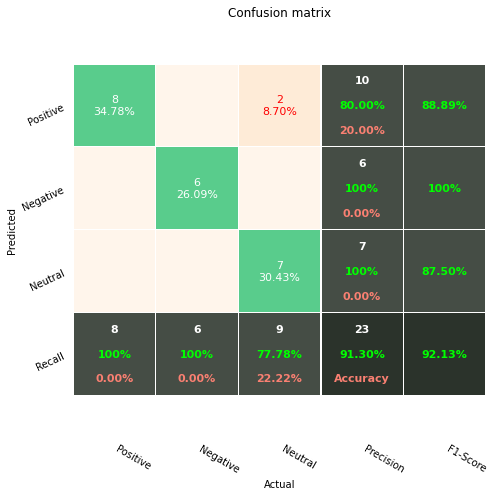}}

\caption{Confusion matrix for GAN-BioBERT }
\label{fig:2}      
\end{figure}

After completion of training, the final GAN-BioBERT algorithm was found to have an accuracy of 91.3\%, and a macro F1-Score of 0.92. The training of the algorithm took forty-five minutes using the Google Colaboratory Environment using 35 GB of RAM with TPU hardware acceleration.  The confusion matrix associated with these results is shown in Figure \ref{fig:2}. 

As a point of comparison, the performance of the standard GAN-BERT approach was assessed as well as the performance of an expert rater on the same dataset. GAN-BERT using the base uncased BERT pretrained model was found to have an accuracy of 82.6\% and a macro F1-score of 0.824. Using the same dataset, an expert rater had an accuracy of 62\% with a macro F1-Score of 0.603. These results, alongside the results of the two previous studies investigating sentiment analysis of clinical trial abstracts, are summarized in Table \ref{tab:1}.

From a technical perspective, these results show that GAN-BioBERT is a significant step forward for assessing the sentiment in clinical trial literature, with an 8.7\% improvement in performance over GAN-BERT for the same classification task. When compared to previous studies' attempts at classifying sentiment in clinical trial abstracts \cite{Zlabinger,Fischer}, this improvement is even more significant as there is an absolute accuracy improvement of 15.3\%, while also expanding the classification task to positive, negative, and neutral, as opposed to positive/not positive \cite{Fischer}, or positive/neutral \cite{Zlabinger}. This significant improvement in accuracy and expansion of the number of classifiers make GAN-BioBERT much more suitable for large-scale assessment of the sentiment in clinical trial literature with improved accuracy and data resolution. With the already high classification accuracy of the algorithm in mind, further development of this algorithm technically may include the introduction of finer-grained sentiment classification, as well as the use of a larger set of labeled training data with more expert raters contributing to improve inference performance.

The performance of the manual expert's rating also merits further discussion of its implications, especially when compared to the performance of GAN-BioBERT. The expert raters used in this study were board-certified physicians with many years working within the setting of academic medicine, and thus are likely to be some of the most qualified individuals to interpret the sentiment in the abstracts of clinical research papers. Knowing this, the expert rater's accuracy of 62\% on this classification task underpins both the difficulty of the task of classifying sentiment in clinical literature, as well as the importance of developing a more accurate and standardized measure to assess this sentiment. This finding also brings into question the validity of the historical practice \cite{Oliveira,Yuan,Vecchi} of large-scale qualitative assessment of clinical literature using manual raters. In short, the comparative performance of expert rating to GAN-BioBERT shows that developing a systematic, reproducible method for assessing clinical literature is a large step forward for the field of academic medicine, and in creating a systematic, unbiased method for assessing the statements made in clinical literature. 

\section{Conclusion}
This study presents GAN-BioBERT, a sentiment analysis classifier for the assessment of the sentiment expressed in clinical trial abstracts. GAN-BioBERT was shown to significantly outperform both previous attempts to classify sentiment in clinical trial abstracts using sentiment analysis as well as manual expert rating with regards to accuracy and number of sentiment classes. Considering this high multi-class accuracy, and the reproducible results GAN-BioBERT generates, this study posits GAN-BioBERT as a viable tool for large-scale assessment of the findings expressed in clinical trial literature. By using a tool such as GAN-BioBERT the large scale assessment of qualitative reporting trends in clinical trial literature becomes significantly more feasible with more reproducible findings when compared to previously utilized methods.

%\begin{acknowledgements}
%\end{acknowledgements}

\section*{Conflict of interest}
The authors declare that they have no conflict of interest.

\end{document}